\documentclass[letterpaper]{article} 
\usepackage{aaai2026}  
\usepackage{times}  
\usepackage{helvet}  
\usepackage{courier}  
\usepackage[hyphens]{url}  
\usepackage{graphicx} 
\usepackage{natbib}  
\usepackage{caption} 
\usepackage{algorithm}
\usepackage{algorithmic}

\usepackage{newfloat}
\usepackage{listings}
\DeclareCaptionStyle{ruled}{labelfont=normalfont,labelsep=colon,strut=off} 
\floatstyle{ruled}
\newfloat{listing}{tb}{lst}{}
\floatname{listing}{Listing}
\title{HiFusion: Hierarchical Intra-Spot Alignment and Regional Context Fusion for Spatial Gene Expression Prediction from Histopathology}
\author{
    Ziqiao Weng\textsuperscript{\rm 1,\rm 4}\thanks{Corresponding author.},
    Yaoyu Fang\textsuperscript{\rm 1},
    Jiahe Qian\textsuperscript{\rm 1,\rm 5},
    Xinkun Wang\textsuperscript{\rm 2},
    Lee AD Cooper\textsuperscript{\rm 3},
    Weidong Cai\textsuperscript{\rm 4},
    Bo Zhou\textsuperscript{\rm 1}\footnotemark[1]
}

\affiliations{
    \textsuperscript{\rm 1}Department of Radiology, Northwestern University, Chicago, 60611, IL, USA.\\
    \textsuperscript{\rm 2}Department of Cell and Developmental Biology, Northwestern University, Chicago, 60611, IL, USA.\\
    \textsuperscript{\rm 3}Department of Pathology, Northwestern University, Chicago, 60611, IL, USA.\\
    \textsuperscript{\rm 4}School of Computer Science, The University of Sydney, Sydney, NSW, 2006, Australia\\
    \textsuperscript{\rm 5}Institute of Automation, Chinese Academy of Sciences, Beijing, 100190, China.\\
    alexziqiaoweng@gmail.com, bo.zhou@northwestern.edu

}

\usepackage{bibentry}

\usepackage{amsfonts} 
\usepackage{amsmath}
\usepackage{caption}
\usepackage{booktabs}
\usepackage{multirow}
\usepackage{makecell}
\usepackage{lscape}   
\usepackage{graphicx}
\usepackage{subcaption} 
\usepackage{subfig} 
\usepackage{mleftright}
\usepackage{xcolor}      
\usepackage{colortbl}    
\usepackage{array} 
\usepackage[skip=4pt]{caption}
\begin{document}
\setlength{\textfloatsep}{5pt}
\maketitle

\begin{abstract}


Spatial transcriptomics (ST) bridges gene expression and tissue morphology but faces clinical adoption barriers due to technical complexity and prohibitive costs. While computational methods predict gene expression from H\&E-stained whole-slide images (WSIs), existing approaches often fail to capture the intricate biological heterogeneity within spots and are susceptible to morphological noise when integrating contextual information from surrounding tissue. To overcome these limitations, we propose HiFusion, a novel deep learning framework that integrates two complementary components. First, we introduce the Hierarchical Intra-Spot Modeling module that extracts fine-grained morphological representations through multi-resolution sub-patch decomposition, guided by a feature alignment loss to ensure semantic consistency across scales. Concurrently, we present the Context-aware Cross-scale Fusion module, which employs cross-attention to selectively incorporate biologically relevant regional context, thereby enhancing representational capacity. This architecture enables comprehensive modeling of both cellular-level features and tissue microenvironmental cues, which are essential for accurate gene expression prediction. Extensive experiments on two benchmark ST datasets demonstrate that HiFusion achieves state-of-the-art performance across both 2D slide-wise cross-validation and more challenging 3D sample-specific scenarios. These results underscore HiFusion’s potential as a robust, accurate, and scalable solution for ST inference from routine histopathology.

\end{abstract}


\begin{links}
    \link{Code}{https://github.com/Advanced-AI-in-Medicine-and-Physics-Lab/HiFusion}
\end{links}

\section{Introduction}\label{sec:introduction}


Spatial transcriptomics (ST) has emerged as a transformative technology that enables genome-wide gene expression profiling while preserving spatial localization within tissue sections, offering near-cellular resolution of molecular activity \cite{zhu2025asign,he2020integrating,pang2021leveraging}. By integrating high-throughput RNA sequencing with spatial barcoding, ST maps transcriptomes to precise histological coordinates, thereby revealing spatial heterogeneity, tissue architecture, and cell–cell interactions across diverse biological systems \cite{he2020integrating,pang2021leveraging}. Most ST platforms divide tissue sections into discrete spots, typically 55–100 µm in diameter \cite{lin2024stalign,niu2025ph2st}. These spatially barcoded spots collectively generate expression matrices that bridge molecular phenotypes with tissue morphology \cite{staahl2016visualization}.

Despite its high-resolution potential, widespread adoption of ST remains hindered by practical constraints, including high experimental costs, specialized instrumentation, and limited scalability in clinical workflows \cite{zhu2025asign,ruiz2025completing,yang2023exemplar}. Consequently, large-scale diagnostic or population-level applications remain rare.

In contrast, hematoxylin and eosin (H\&E)-stained whole-slide images (WSIs) are routinely acquired in clinical pathology, are cost-effective, and encapsulate rich morphological features closely associated with gene expression patterns \cite{zhu2025asign,niu2025ph2st}. For example, overexpression of tumor markers such as \textit{ERBB2} in HER2-positive breast cancer has been linked to distinct morphological phenotypes in H\&E images \cite{NEURIPS2024_3ef2b740}. This has motivated a growing body of research on computational models, particularly deep learning-based approaches, to infer transcriptomic profiles directly from WSIs \cite{he2020integrating,pang2021leveraging}.

Recent advances in deep neural networks, including convolutional neural networks (CNNs), graph neural networks (GNNs), and transformer-based architectures, have enabled the prediction of spatially resolved gene expression at the spot level directly from WSI-derived image patches \cite{zhu2025asign,xie2023spatially}. These models typically take spot-aligned image patches as input and aim to predict the expression levels of hundreds to thousands of genes by learning complex associations between tissue morphology and molecular profiles \cite{NEURIPS2024_3ef2b740,pang2021leveraging}. For instance, ST-Net \cite{he2020integrating} employs a DenseNet backbone to generate spot-level predictions, while more recent approaches, such as HisToGene, Hist2ST, EGN, TRIPLEX, and ASIGN, enhance performance by incorporating spatial dependencies via long-range modeling, multi-resolution inputs, and inter-spot context integration \cite{pang2021leveraging,zeng2022spatial,chung2024accurate,zhu2025asign}.

Despite these advances, several key limitations remain. First, most existing methods struggle to capture both fine-grained morphological details and global tissue context simultaneously \cite{chen2025delstdualentailmentlearning,chung2024accurate}. They typically treat each spot as homogeneous, overlooking the hierarchical structure within the spot. In reality, a single spot often contains diverse microstructures, such as distinct cell types, nuclear textures, and subcellular patterns, that are directly associated with gene expression \cite{pang2021leveraging,he2020integrating}. However, current architectures often fail to exploit these multi-scale intra-spot cues.

Moreover, broader contextual information is often used merely as auxiliary input, without explicitly modeling the semantic correlations between a spot and its surrounding tissue. This decoupled design limits the effective integration of region-aware signals, potentially leading to suboptimal representations for spatial gene prediction. Although recent models like TRIPLEX and ASIGN have adopted large regional patches (e.g., exceeding 1000×1000 pixels) to incorporate spatial context, the utility of such high-resolution inputs remains empirically underexplored. In fact, expanding the receptive field may introduce morphological noise or irrelevant signals, especially when adjacent regions lack biological relevance to the target spot.

To address these challenges, we propose \textbf{\textit{HiFusion}} (Hierarchical Intra-Spot Alignment and Context-aware Fusion Network), a dual-branch framework for robust spatial gene expression prediction from histopathology images. HiFusion comprises two key components: Hierarchical Intra-Spot Modeling (HISM) and Context-Aware Cross-Scale Fusion (CCF).

The HISM module explicitly captures intra-spot morphological heterogeneity by decomposing each spot image into a hierarchy of non-overlapping sub-patches at multiple spatial resolutions, down to the cellular level. These multi-scale patches, along with the full-spot image, are processed through a shared encoder to extract features reflecting tissue-, cellular-, and subcellular-level structures. A feature alignment loss ensures semantic consistency across scales, encouraging coherent multi-scale representations.

The CCF module incorporates broader tissue context by encoding neighboring regions with a lightweight encoder. Contextual features act as queries in a cross-attention module, while the adaptively fused multi-scale spot representations from HISM serve as keys and values. This design allows the model to selectively attend to biologically relevant contextual information while suppressing spatial noise, thereby enhancing the robustness and expressiveness of the learned features.


Together, these two components enable HiFusion to jointly model fine-grained intra-spot morphology and spatial context, overcoming limitations of coarse spot-level representations and simplistic context integration. 

Our key contributions can be summarized as follows:

\begin{itemize}
    \item We propose HiFusion, a novel framework for spatial gene expression prediction from whole-slide images. HiFusion explicitly integrates multi-scale intra-spot representations with regional tissue context, effectively capturing spatial and biological heterogeneity across scales.

    \item Our approach introduces a hierarchical intra-spot modeling module that extracts rich, fine-grained features from multiple spatial resolutions, coupled with a feature alignment loss to ensure semantic consistency across scales. A context-aware cross-scale fusion module further integrates these intra-spot features with neighboring regional context via a residual cross-attention mechanism, enhancing representational expressiveness and robustness.

    \item Extensive evaluation on two public ST datasets demonstrates that HiFusion consistently outperforms state-of-the-art methods under both conventional 2D slide-wise cross-validation and a recent 3D sample-specific evaluation protocol, establishing new benchmarks for spatial gene expression inference. Comprehensive ablation studies further validate the effectiveness of each module and analyze the impact of spatial context size on prediction performance.
\end{itemize}

\section{Related Works}\label{sec:related_works}

Spatial transcriptomics captures spatially resolved mRNA using microarray chips, followed by next-generation sequencing and spatial mapping onto histological images to generate high-resolution gene expression landscapes \cite{zhang2022clinical,niu2025ph2st}. Recent efforts have shifted toward learning-based approaches that infer spatial gene expression directly from H\&E-stained WSIs, formulating it as a multi-output regression task over spot-aligned image patches.

ST-Net \cite{he2020integrating} initiated this direction by mapping spot-level patches to gene expression using a DenseNet-121 backbone, treating each spot independently and neglecting contextual cues. HisToGene \cite{pang2021leveraging} incorporated long-range dependencies via Vision Transformers, while Hist2ST \cite{zeng2022spatial} added local feature extraction (ConvMixer) and neighborhood modeling with Graph Neural Networks. Image similarity–based models like EGN \cite{yang2023exemplar} and BLEEP \cite{xie2023spatially} retrieve exemplar patches or learn contrastive embeddings, but are sensitive to staining variations and generalize poorly across samples. To integrate broader context, TRIPLEX \cite{chung2024accurate} extracts features from the spot, its neighborhood, and the full slide using a three-branch architecture, while ASIGN \cite{zhu2025asign} aligns adjacent tissue sections in 3D with a graph-based model. However, both methods rely on multi-resolution inputs and complex alignment pipelines. In contrast, our proposed HiFusion explicitly captures intra-spot hierarchical structure via multiscale patch decomposition and enforces cross-scale consistency through feature alignment. It further enhances context integration by fusing region-level features with fine-grained spot representations via cross-attention, offering a more efficient and generalizable framework for spatial gene expression prediction.
\section{Method}\label{sec:method}

\subsection{Problem Formulation}

\begin{figure*}[t]
\centering
\includegraphics[width=\linewidth]{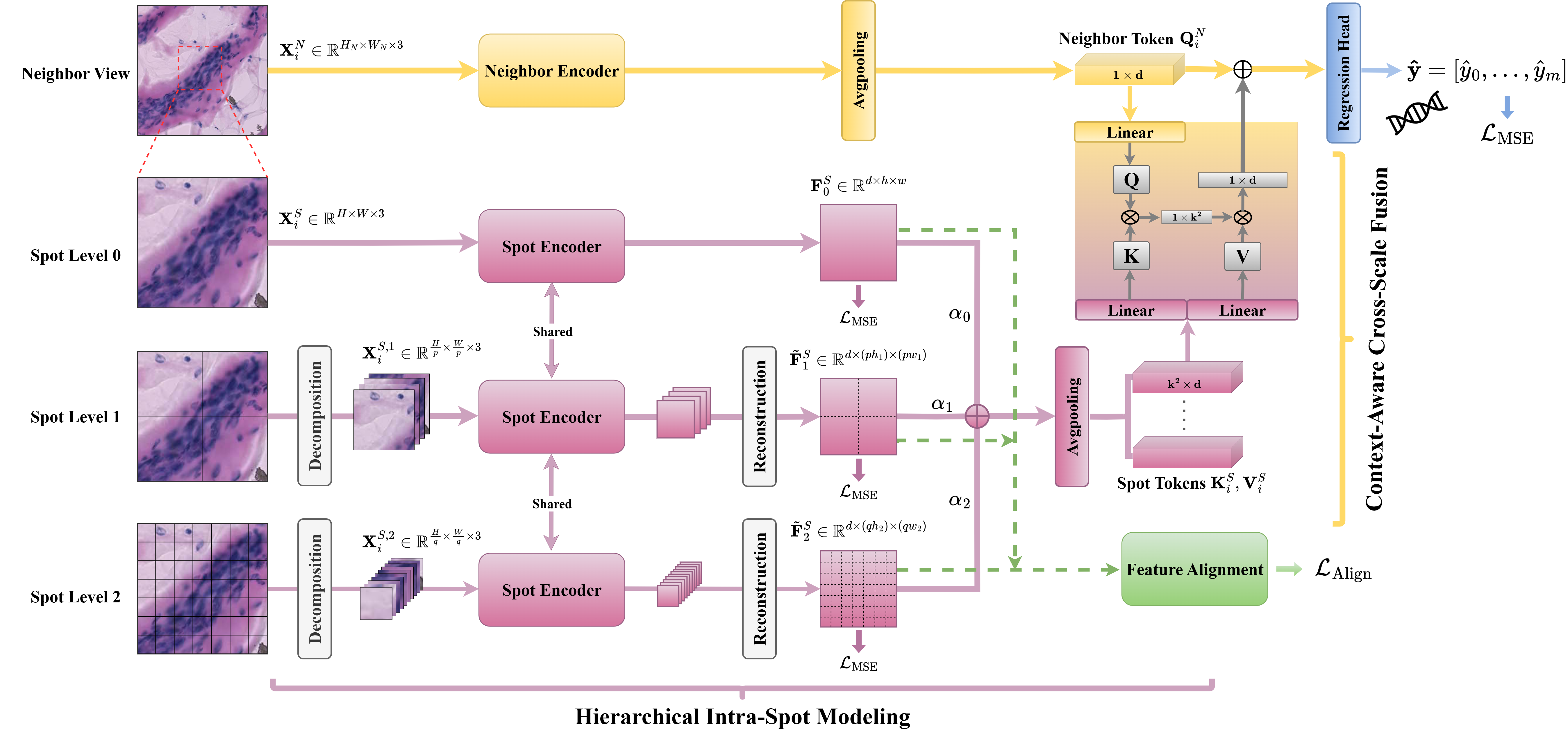} 
\caption{Schematic of the proposed \textbf{HiFusion} framework, which integrates \textit{Hierarchical Intra-Spot Modeling (HISM)} and \textit{Context-Aware Cross-Scale Fusion (CCF)}. HISM hierarchically decomposes each spot into multi-scale patches to extract fine-grained features with semantic alignment. CCF fuses contextual region features with multi-scale spot representations via residual cross-attention for gene expression prediction.}
\label{fig:fig1}
\end{figure*}

We formulate spatial gene expression prediction as a multi-output regression task over a set of spatially arranged spots on a whole-slide image (WSI). Let $X^{S} \in \mathbb{R}^{n \times H_S \times W_S \times 3}$ denote the collection of cropped spot-level image patches, where $n$ is the number of spots, and $(H_S, W_S)$ denotes the height and width of each spot image. The corresponding normalized gene expression profiles are represented as $Y \in \mathbb{R}^{n \times m}$, where $m$ is the number of genes to be predicted. To incorporate local spatial context, we additionally extract regional neighbor patches for each spot, denoted as $X^{N} \in \mathbb{R}^{n \times H_N \times W_N \times 3}$, where $(H_N, W_N)$ denotes the size of each neighbor patch. Our objective is to learn a predictive function $\phi: \{X^{S}, X^{N}\} \rightarrow Y$ that maps each spot and its surrounding context to its corresponding gene expression vector.

\subsection{Overview of HiFusion}

The overall workflow of the proposed framework, HiFusion, is illustrated in Figure \ref{fig:fig1}. It consists of two main components: Hierarchical Intra-Spot Modeling (HISM) and Context-Aware Cross-Scale Fusion (CCF). In HISM, each spot image is decomposed into non-overlapping sub-patches at multiple resolutions, including the cellular scale, to capture fine-grained morphological cues such as nuclear structure and cell-type variation. The full spot and its sub-patches are processed by a shared encoder to extract multi-scale features. A feature alignment loss encourages semantic consistency across scales, leveraging the translation invariance of CNNs. In CCF, multi-scale spot features are adaptively fused to form the keys and values, while a region-level feature extracted by a lightweight encoder serves as the query. These are integrated via residual multi-head cross-attention to produce the final representation used for gene expression prediction.

\subsection{Hierarchical Intra-Spot Modeling and Alignment}

To capture the rich intra-spot morphological heterogeneity, we propose a hierarchical modeling strategy that encodes visual patterns from tissue- to subcellular-level resolution. Given a spot image $\mathbf{X}_i^{S} \in \mathbb{R}^{H \times W \times 3}$, we define this as the Level-0 input. A shared encoder $f_\theta(\cdot)$ extracts a global feature map $\mathbf{F}_0^{S} = f_\theta(\mathbf{X}_i^{S}) \in \mathbb{R}^{d \times h \times w}$, which captures coarse tissue-level context.

To obtain finer-scale features, we decompose the input into $p \times p$ and $q \times q$ non-overlapping patches ($q > p$), forming Level-1 and Level-2 inputs $\{\mathbf{X}_{i,j}^{S,1}\}_{j=1}^{p^2}$ and $\{\mathbf{X}_{i,k}^{S,2}\}_{k=1}^{q^2}$, where $\mathbf{X}_{i,j}^{S,1} \in \mathbb{R}^{\frac{H}{p} \times \frac{W}{p} \times 3}$ and $\mathbf{X}_{i,k}^{S,2} \in \mathbb{R}^{\frac{H}{q} \times \frac{W}{q} \times 3}$, respectively. These patches are passed through the same encoder to yield multi-scale representations: $\mathbf{F}_1^{S} \in \mathbb{R}^{p^2 \times d \times h_1 \times w_1}$ and $\mathbf{F}_2^{S} \in \mathbb{R}^{q^2 \times d \times h_2 \times w_2}$.

We then reconstruct the spatial layout of patch features based on their original positions, resulting in $\tilde{\mathbf{F}}_1^{S} \in \mathbb{R}^{d \times (p h_1) \times (p w_1)}$ and $\tilde{\mathbf{F}}_2^{S} \in \mathbb{R}^{d \times (q h_2) \times (q w_2)}$. If the reconstructed resolutions do not match that of $\mathbf{F}_0^{S}$, bilinear interpolation is applied to align them accordingly.

To enforce cross-scale semantic consistency, we define a feature alignment loss that encourages the fine-scale features to preserve the global semantics of the full spot representation:

\begin{equation}
\mathcal{L}_{\text{align}} = \sum_{s=1}^{2} \left\| \tilde{\mathbf{F}}_s^{S} - \mathbf{F}_0^{S} \right\|_1
\end{equation}

This hierarchical design enables the model to learn both coarse and fine-grained morphological representations within each spot. By explicitly aligning multi-scale features at the pixel level, the network is encouraged to maintain semantic consistency across spatial resolutions, thereby enhancing the robustness of the learned representations.



\subsection{Context-Aware Cross-Scale Fusion}

To incorporate broader tissue context while preserving intra-spot structural fidelity, we introduce a cross-scale fusion module guided by region-level information. For each spot $\mathbf{X}_i^{S}$, we extract a surrounding tissue region $\mathbf{X}_i^{N} \in \mathbb{R}^{H_{N} \times W_{N} \times 3}$, which is encoded by a lightweight encoder $f_\psi(\cdot)$ followed by global average pooling, yielding a condensed regional representation $\mathbf{Q}_i^{N} \in \mathbb{R}^{1 \times d}$:

\begin{equation}
\mathbf{Q}_i^{N} = \text{AvgPool}(f_\psi(\mathbf{X}_i^{N}))
\end{equation}

To integrate multi-scale intra-spot features \(\{\mathbf{F}_s^{S}\}_{s=0}^{2}\) derived from spot image decomposition at three scales (\(s=0,1,2\)), we adopt a learnable weighted fusion strategy. The fused spot representation is computed as:

\begin{equation}
\mathbf{F}_{\text{fused}}^{S} = \sum_{s=0}^{2} \omega_s \cdot \mathbf{F}_s^{S} ,
\end{equation}
where the weight \(\omega_s\) for each scale is generated via a softmax over learnable parameters \(\{\alpha_s\}_{s=0}^{2}\):

\begin{equation}
\omega_s = \frac{\exp(\alpha_s)}{\sum_{j=0}^{2} \exp(\alpha_j)}, \quad s \in \{0,1,2\}
\end{equation}
To prepare the fused intra-spot features for cross-attention, we apply adaptive average pooling to obtain \(\bar{\mathbf{F}}^{S}=\text{AvgPool}(\mathbf{F}_{\text{fused}}^{S},(k,k)) \in \mathbb{R}^{d \times k \times k}\), which is then reshaped into key and value matrices:
\begin{equation}
\mathbf{K}_i^{S} = \mathbf{V}_i^{S} = \text{reshape}(\bar{\mathbf{F}}^{S}) \in \mathbb{R}^{k^2 \times d}
\end{equation}


We then integrate the region-level and multi-scale intra-spot features through a residual cross-attention mechanism. Specifically, the region token $\mathbf{Q}^{N}$ is used as the query, while the fused intra-spot representation serves as the keys and values. The cross-attention output is computed as:

\begin{equation}
\phi_{\text{ca}}(\cdot) = \text{softmax}\left( \frac{(\mathbf{Q}_i^{N} \mathbf{W}_Q)(\mathbf{K}_i^{S} \mathbf{W}_K)^\top}{\sqrt{d_K}} \right)(\mathbf{V}_i^{S} \mathbf{W}_V),
\end{equation} 
where $\mathbf{W}_Q$, $\mathbf{W}_K$, and $\mathbf{W}_V$ are learnable linear projections, and $d_K$ is the dimensionality of the key vectors used for scaling. Here we present the single-head formulation; our implementation uses the standard multi-head attention.

Finally, we predict the gene expression vector $\hat{\mathbf{y}}_i \in \mathbb{R}^m$ using the fused output and a residual connection, followed by a prediction head composed of LayerNorm and a fully connected layer:
\begin{equation}
\hat{\mathbf{y}}_i = \text{FC}(\text{LayerNorm}(\mathbf{Q}_i^{N} + \phi_{\text{ca}}(\mathbf{Q}_i^{N}, \mathbf{K}_i^{S}, \mathbf{V}_i^{S})))
\end{equation}

This context-aware fusion strategy allows the model to selectively attend to semantically relevant intra-spot features while mitigating high-frequency noise introduced by overly fine-grained morphological details.

\subsection{Loss Function}

To enhance context-aware gene expression prediction, we employ a composite loss combining regression objectives with feature alignment.

Let $\hat{\mathbf{y}}_i, \mathbf{y}_i \in \mathbb{R}^m$ denote the predicted and ground-truth expression for the $i$-th spot. The primary prediction loss is:
\begin{equation}
\mathcal{L}_{\text{main}} = \frac{1}{n} \sum_{i=1}^n \| \hat{\mathbf{y}}_i - \mathbf{y}_i \|_2^2
\end{equation}

For multi-scale supervision, each level's features ($\mathbf{F}_0^{S}$, $\tilde{\mathbf{F}}_1^{S}$, $\tilde{\mathbf{F}}_2^{S}$) generate auxiliary predictions $\hat{\mathbf{y}}_i^{(s)}$ via shared FC layers:

\begin{equation}
\mathcal{L}_{\text{aux}} = \frac{1}{3n} \sum_{i=1}^n \sum_{s=0}^2 \| \hat{\mathbf{y}}_i^{(s)} - \mathbf{y}_i \|_2^2
\end{equation}

The total training objective combines regression and alignment losses:

\begin{equation}
\mathcal{L}_{\text{total}} = \underbrace{\mathcal{L}_{\text{main}} + \mathcal{L}_{\text{aux}}}_{\mathcal{L}_{\text{reg}}} + \lambda \mathcal{L}_{\text{align}},
\end{equation}
where $\lambda$ balances the alignment regularization. This multi-level supervision strategy ensures that the network captures both global and fine-grained morphological patterns relevant to gene expression, while the alignment term enforces semantic consistency across scales to improve training stability and generalization. 

\section{Experiments and Results}\label{sec:experiments}

\subsection{Dataset and Pre-processing}
In our experiments, we utilized two publicly available datasets used in \cite{zhu2025asign} to evaluate the performance of the proposed \textit{HiFusion} model. The first is the HER2-positive breast tumor dataset \cite{andersson2021spatial}, denoted as HER2, which includes 36 whole-slide images comprising 4 samples with 6-layer tissue sections and 4 samples with 3-layer sections, totaling 13,620 ST spots. The second dataset, referred to as ST-Data, is a breast cancer dataset introduced in ST-Net \cite{he2020integrating}. Following the protocol in \cite{zhu2025asign}, we selected 16 samples with three-layer tissue sections, yielding 41,544 spots in total. In both datasets, the spots have a diameter of 100~\textmu m and are arranged on a grid with a center-to-center distance of 200~\textmu m.


Given the high dimensionality of gene expression data (exceeding 15,000 genes), it is impractical to predict the expression of all genes from histological patches. Hence, following \cite{he2020integrating}, we select the top 250 genes with the highest average expression levels for prediction. The selected genes are listed in the supplementary material. To normalize gene expression values, we first perform a spot-wise normalization by dividing each gene count $x_i$ by the total expression count across all genes within the same spot (after adding 1 to avoid division by zero), and then apply a logarithmic transformation $x^{\text{norm}} = \log\left((x + 1)/\sum_{i=1}^{m}(x_i + 1)\right)$, where $m$ denotes the number of genes and all operations are applied element-wise. 


\subsection{Experiment Setup and Evaluation Metrics}




We evaluate our method under two distinct testing scenarios to comprehensively assess its performance.

The first, 2D slide-wise cross-validation, follows the protocol commonly adopted in prior studies. We conduct 4-fold cross-validation on both datasets, ensuring that samples from the same patient are assigned exclusively to either the training or test set to avoid data leakage.

The second, 3D sample-specific validation, represents a more challenging and recently proposed paradigm \cite{qian2025stdaisingleshot25dspatial,fang2025}, where training and testing are performed within individual patients. Specifically, the first histological layer from each patient is used for training, while the remaining sections are reserved for testing. This setup emphasizes within-sample generalization and minimizes inter-patient domain shifts.

Model performance is evaluated using three metrics: Mean Squared Error (MSE), which reflects the average squared deviation between predictions and ground truth; Mean Absolute Error (MAE), capturing the average magnitude of errors; and the Pearson Correlation Coefficient (PCC), which quantifies the linear correlation between predicted and actual gene expression values. Lower MSE and MAE indicate higher predictive accuracy, while higher PCC suggests stronger consistency with ground-truth profiles.

\subsection{Implementation Details}
For all datasets, each spot image (Level-0 input) is cropped to $224 \times 224$ pixels, corresponding to approximately $150~\mu$m $\times$ $150~\mu$m in the original pathology image, using the center coordinates of each spot. To model intra-spot spatial hierarchy, each spot image is decomposed into $2 \times 2$ and $7 \times 7$ non-overlapping patches (Level-1 and Level-2 inputs, respectively), where $p=2$ and $q=7$. To incorporate regional context, neighboring patches are extracted by cropping a $448 \times 448$ image centered at each spot. We adopt ResNet-18 as the backbone encoder for spot-level feature extraction and ResNet-10 as a lightweight encoder for neighboring regions. The number of key and value tokens after adaptive average pooling is set to $k \times k = 2 \times 2$. The loss weight $\lambda$ is empirically set to 1. The model is optimized using Adam with a momentum of 0.9 and a weight decay of $10^{-5}$. The initial learning rate is set to $3 \times 10^{-4}$ and adjusted dynamically using a cosine annealing scheduler with a minimum learning rate ($\eta_{\text{min}}$) of $1 \times 10^{-6}$. Training is conducted for 50 epochs with a batch size of 32. The reported results are the average performance across all patient samples. All experiments are performed on a single NVIDIA RTX 4090 GPU.

\subsection{Baselines}
We benchmarked the performance of our model against several representative baseline methods, including: 1) local-based models (ST-Net~\cite{he2020integrating}, EGN~\cite{yang2023exemplar}) and 2) global-based models (TRIPLEX~\cite{chung2024accurate}, ASIGN~\cite{zhu2025asign}). Notably, ASIGN represents the current state-of-the-art (SOTA) approach. We reproduced ST-Net and EGN following the implementations described in the TRIPLEX and ASIGN papers. Specifically, for EGN, we employed ResNet-18 as the feature extractor, while ST-Net utilized a pretrained DenseNet-121. For TRIPLEX and ASIGN, we retained the original network architectures and hyperparameter settings as reported in their respective papers. All baseline models were trained and evaluated under identical conditions to ensure a fair comparison.

\subsection{Comparison between HiFusion and Baselines}

\begin{table*}[]
\centering
\setlength{\tabcolsep}{5pt}
\label{tab:main_results}
\small
\begin{tabular}{c|ccc|ccc||ccc|ccc}
\toprule
\multirow{3}{*}{\textbf{Method}} 
& \multicolumn{6}{c||}{\textbf{HER2}} 
& \multicolumn{6}{c}{\textbf{ST-Data}} \\
\cmidrule(lr){2-7} \cmidrule(lr){8-13}
& \multicolumn{3}{c|}{2D slide-wise} 
& \multicolumn{3}{c||}{3D sample-specific} 
& \multicolumn{3}{c|}{2D slide-wise} 
& \multicolumn{3}{c}{3D sample-specific} \\
\cmidrule(lr){2-4} \cmidrule(lr){5-7} \cmidrule(lr){8-10} \cmidrule(lr){11-13}
& MSE & MAE & PCC & MSE & MAE & PCC & MSE & MAE & PCC & MSE & MAE & PCC \\
\midrule
ST-Net  & 0.6523 & 0.6255 & 0.4621 & 0.5323 & 0.5747 & 0.7042 & 0.5798 & 0.5943 & 0.5304 & 0.4939 & 0.5514 & 0.7443 \\

HisToGene & 0.6105 & 0.6063 & 0.4294 & 0.4851 & 0.4899 & 0.7028 & 0.5310 & 0.5694 & 0.5427 & 0.4841 & 0.5121 & 0.7280 \\

His2ST & 0.5843 & \underline{0.5885} & 0.4478 & 0.3174 & 0.4438 & 0.7276 & \underline{0.5230} & \underline{0.5636} & \underline{0.5442} & 0.2877 & 0.4240 & 0.7725 \\

EGN & 0.5845 & 0.5940 & 0.4723 & 0.2917 & \underline{0.4258} & 0.7441 & 0.5568 & 0.5800 & 0.5103 & \underline{0.2755} & \underline{0.4136} & \underline{0.7800} \\

TRIPLEX & \underline{0.5715} & 0.5918 & \underline{0.4750} & \underline{0.2899} & 0.4268 & \underline{0.7471} & 0.5389 & 0.5769 & 0.5387 & 0.2857 & 0.4255 & 0.7780 \\

ASIGN-2D & 0.5830 & 0.5901 & 0.4601 & 0.3116 & 0.4415 & 0.7316 & 0.5449 & 0.5764 & 0.5373 & 0.2822 & 0.4204 & 0.7741 \\

ASIGN-3D  & \textbackslash & \textbackslash & \textbackslash & 0.4163
 & 0.4987 & 0.7019
 & \textbackslash & \textbackslash & \textbackslash & 0.3141 & 0.4445 & 0.7524 \\

\rowcolor{gray!20} \textbf{HiFusion$^{*}$ (Ours)} & \textbf{0.5459} & \textbf{0.5699} & \textbf{0.4961} & \textbf{0.2846} & \textbf{0.4205} & \textbf{0.7492} & \textbf{0.5095} & \textbf{0.5557} & \textbf{0.5613} & \textbf{0.2711} & \textbf{0.4102} & \textbf{0.7838} \\
\bottomrule
\end{tabular}
\caption{Comparison of MSE, MAE, and PCC on HER2 and ST-Data under two evaluation settings. Best and second-best results are shown in bold and underlined, respectively. $^{*}$ denotes significant improvement over the second-best baseline ($p<0.05$).}
\label{tab:tab1}
\end{table*}


Table~\ref{tab:tab1} demonstrates the superior performance of HiFusion across both HER2 and ST-Data datasets under 2D slide-wise and 3D sample-specific evaluations. In slide-wise testing, HiFusion achieved MSE/MAE/PCC scores of 0.5459/0.5699/0.4961 on HER2, outperforming TRIPLEX (second-best) by 2.1–2.6\% and ASIGN (SOTA) by 2.0–3.7\%, with over 10\% MSE improvement versus ST-Net. Similar superiority was observed on ST-Data, confirming HiFusion's robust cross-patient generalization capability for heterogeneous spatial transcriptomics. The 3D sample-specific evaluation revealed consistent advantages, particularly showing 22–25\% improvement over ST-Net. While less pronounced than 2D slide-wise gains, these results validate HiFusion's effectiveness in modeling intra-patient spatial structures. The consistent superiority across both evaluation paradigms highlights HiFusion's versatility in addressing distinct challenges in ST analysis.

\subsection{Comparative Analysis of 3D Prediction Strategies}

Notably, ASIGN-3D was specifically designed for 3D sample-wise prediction by leveraging known-label layers to improve accuracy in unlabeled sections. However, this sophisticated strategy underperformed compared to the simpler intra-sample learning paradigm, which trains solely on a single labeled section and directly predicts gene expression in adjacent layers. We attribute this unexpected outcome to two key factors: (1) substantial inter-patient variability in histopathology and gene expression, which introduces noise during multi-sample training; and (2) potential feature degradation caused by global 3D registeration during preprocessing, which may distort biologically meaningful tissue structures.

These findings highlight the advantages of the 3D intra-sample learning strategy, which offers both (i) improved prediction accuracy by focusing on patient-specific patterns and (ii) reduced computational burden. This balance of performance and efficiency makes it particularly promising for clinical applications of spatial transcriptomics. Additional comparisons between the 3D and 2D learning paradigms are provided in the supplementary material.

\begin{table}[t]
\centering
\begin{tabular}{l|ccc}
\toprule
\textbf{Level Combination} & \textbf{MSE} $\downarrow$ & \textbf{MAE} $\downarrow$ & \textbf{PCC} $\uparrow$ \\
\midrule
1×1                       & 0.5718 & 0.5840 & 0.4763 \\
1×1 + 2×2                 & 0.5561 & 0.5774 & 0.4902 \\
1×1 + 4×4                 & 0.5538 & 0.5769 & 0.4854 \\
1×1 + 7×7                 & 0.5541 & 0.5725 & 0.4777 \\
1×1 + 2×2 + 4×4           & 0.5478 & 0.5732 & 0.4935 \\
1×1 + 2×2 + 7×7           & \textbf{0.5459} & \textbf{0.5699} & \textbf{0.4961} \\
1×1 + 4×4 + 7×7           & 0.5566 & 0.5765 & 0.4756 \\
1×1 + 2×2 + 4×4 + 7×7     & 0.5477 & 0.5719 & 0.4871 \\
\bottomrule
\end{tabular}
\caption{Ablation study for image decomposition levels}
\label{tab:tab2}
\end{table}



\begin{table}[t]
\centering
\small
\label{tab:ablation_fa}
\setlength{\tabcolsep}{3pt} 
\begin{tabular}{c|c c c|c c c}
\hline
\multirow{2}{*}{\textbf{Method}} & \multicolumn{3}{c|}{\textbf{HER2}} & \multicolumn{3}{c}{\textbf{ST}} \\
\cmidrule(lr){2-4} \cmidrule(lr){5-7} 
& MSE & MAE & PCC & MSE & MAE & PCC \\
\midrule
Our & \textbf{0.5459} & \textbf{0.5699} & \textbf{0.4961} 
    & \textbf{0.5095} & \textbf{0.5557} & \textbf{0.5613} \\
w/o FA & 0.5642 & 0.5798 & 0.4730 
      & 0.5150 & 0.5594 & 0.5559 \\
\hline
\end{tabular}
\caption{Ablation study for feature alignment}
\label{tab:tab3}
\end{table}
\begin{figure}[t]
  \centering

  \begin{subfigure}[b]{0.75\linewidth}
    \centering
    \includegraphics[width=\linewidth]{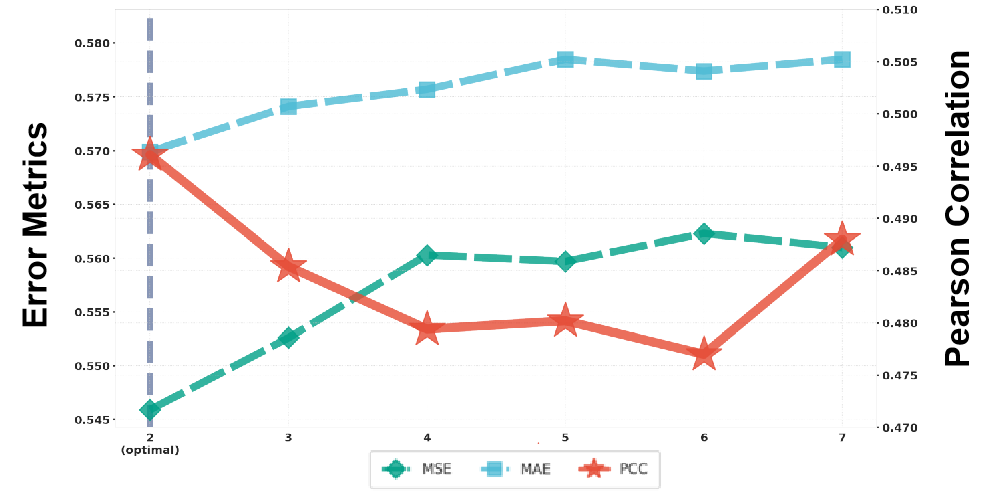}
    \caption{Spot Token Number $(k,k)$}
    \label{fig:token_num}
  \end{subfigure}

  \begin{subfigure}[b]{0.75\linewidth}
    \centering
    \includegraphics[width=\linewidth]{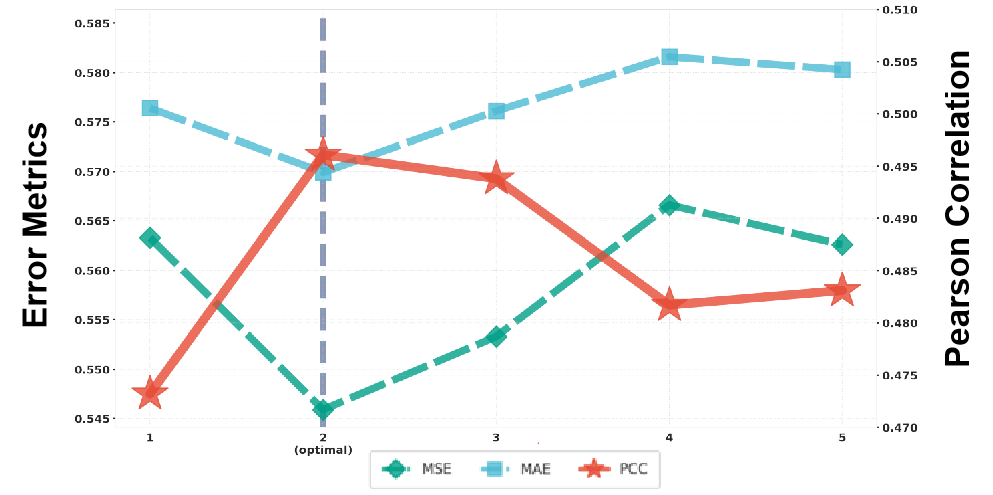}
    \caption{Neighbor Patch Size $(224 \times N,224 \times N)$}
    \label{fig:neighbor_size}
  \end{subfigure}

  \caption{Ablation study for (a) spot token number and (b) neighbor patch size.}
  \label{fig:fig2}
\end{figure}

\subsection{Ablation Study on HISM}

We evaluate the impact of different image decomposition level combinations within our HISM framework on the HER2 dataset. Given the original spot image size of $224 \times 224$, it can be decomposed into patch grids of $2 \times 2$, $4 \times 4$, and $7 \times 7$. Here, $1 \times 1$ refers to using the original image without decomposition. As shown in Table~\ref{tab:tab2}, incorporating any level of decomposition consistently improves performance compared to using only the original image. Even without decomposition, the result is already comparable to the second-best baseline (e.g., TRIPLEX in Table~1), highlighting the strength of our base framework. Among all configurations, the combination of $1 \times 1 + 2 \times 2 + 7 \times 7$ achieves the best overall performance. We attribute this to the complementary spatial granularity captured at different scales: the $1 \times 1$ input preserves global tissue-level context, the $2 \times 2$ patches capture sub-tissue or regional structures, and the fine-grained $7 \times 7$ decomposition provides detailed cellular or subcellular information. This multi-scale representation supports more accurate and biologically meaningful gene expression prediction.

We further investigate the role of feature alignment in HISM on both the HER2 and ST-Data datasets. As shown in Table~\ref{tab:tab3}, incorporating the feature alignment loss consistently improves performance across both datasets. Notably, on HER2, it achieves nearly a 2\% reduction in MSE and over a 2\% increase in PCC. These results suggest that the feature alignment loss effectively enforces cross-scale semantic and predictive consistency, working in synergy with hierarchical image decomposition to enhance overall model performance.

\setlength{\textfloatsep}{10pt}
\begin{figure*}[t]
  \centering
  \includegraphics[width=\textwidth]{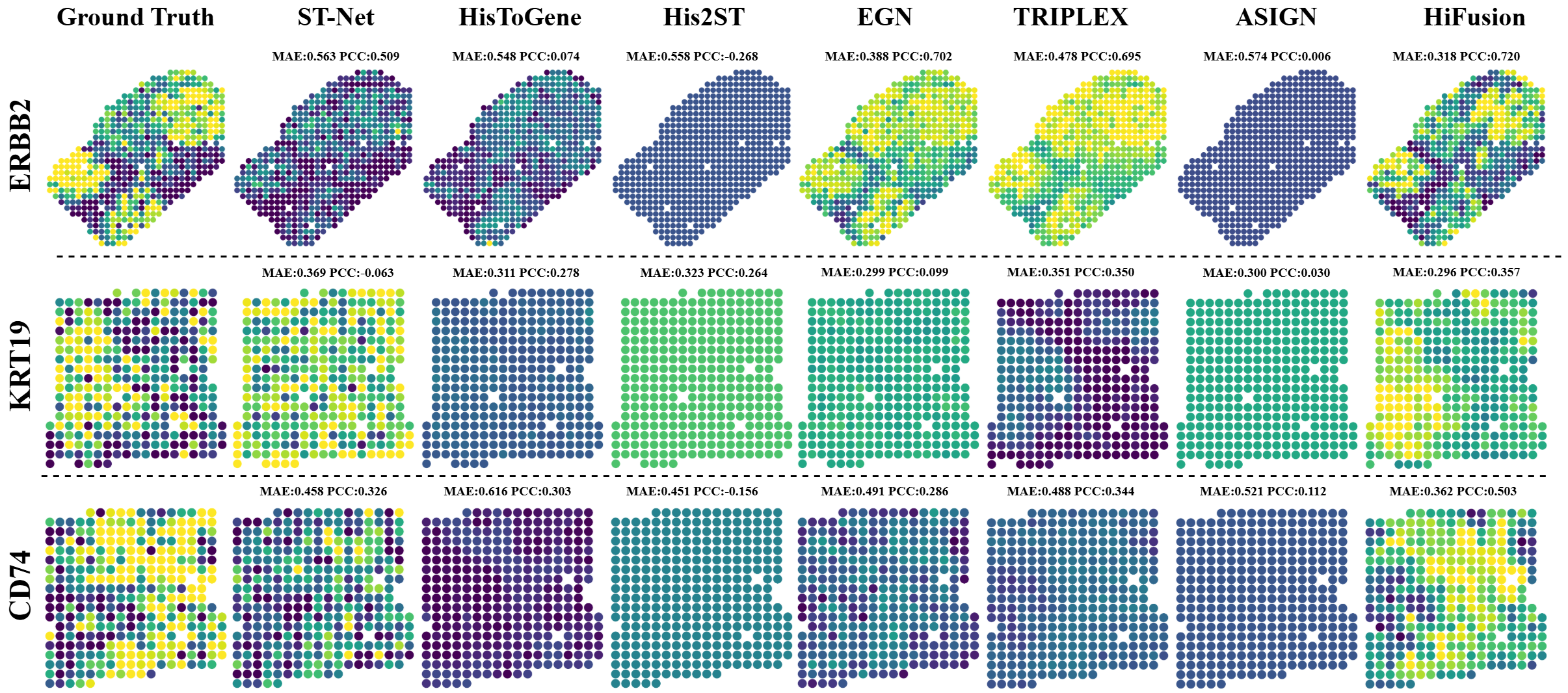} 
  \caption{Predicted spatial expression of ERBB2, KRT19 and CD74 by different models. MAE, PCC values with the ground truth are shown. Brighter regions indicate higher gene expression levels, while darker regions represent lower expression. HiFusion achieves the best visual and quantitative alignment.}
  \label{fig:fig3}
\end{figure*}

\subsection{Ablation Study on CCF}









We conduct two experiments on the HER2 dataset to evaluate the effect of \textbf{(a)} spot token number and \textbf{(b)} neighbor image size in the CCF module, as shown in Figure~\ref{fig:fig2}.

In Figure~\ref{fig:fig2}(a), we vary the number of tokens by applying adaptive average pooling to the fused multi-scale spot features, using grid sizes from $2 \times 2$ to $7 \times 7$ ($7 \times 7$ corresponds to no pooling). Overall, MSE and MAE tend to increase and PCC slightly declines as token count grows. The best performance is observed with the $2 \times 2$ configuration. We hypothesize that fewer tokens help suppress spatial noise while preserving key contextual signals, and are better suited to the capacity of the cross-attention module. In contrast, longer sequences may fragment attention, introduce redundancy, or exceed the module's effective modeling range, leading to performance degradation.

As discussed in the Introduction, prior methods such as TRIPLEX and ASIGN use large patches (e.g., $1120 \times 1120$) to incorporate global context, though their utility remains empirically unclear. To investigate this, Figure~\ref{fig:fig2}(b) shows results using different neighbor image sizes. The optimal performance occurs when the neighborhood is twice the spot size ($N=2$), while larger contexts lead to degradation. This suggests that moderate expansion captures informative spatial cues, whereas excessive enlargement introduces irrelevant tissue regions, increasing noise and diminishing neighbor query quality. Further ablation results on ST-Data are in the supplementary material.

\subsection{Visualization of Cancer Marker Genes}


 To evaluate the clinical applicability of our model, we examined three well-established cancer marker genes with significant relevance to HER2-positive breast cancer: \textit{ERBB2} (HER2) \citep{mehta2014co,dent2013her2,conley2016her2}, \textit{KRT19} \citep{saha2017krt19,tang2014novel}, and \textit{CD74} \citep{su2017biological,borghese2011cd74}. Cross-validation on the HER2 dataset shows that HiFusion consistently outperforms existing methods in predicting spatial gene expression. For \textit{ERBB2}, HiFusion achieved an MAE of 0.711 (PCC = 0.518), substantially outperforming ASIGN (1.074, PCC = $-$0.035), TRIPLEX (0.900, 0.486), EGN (0.778, 0.401), His2ST (1.265, $-$0.029), HisToGene (1.072, 0.267), and ST-Net (1.127, 0.378). Similar performance gains were observed for KRT19, where HiFusion obtained an MAE of 0.446 and PCC of 0.230, surpassing the best competing MAE of 0.450 and PCC of 0.201. For CD74, HiFusion again led with an MAE of 0.584 and PCC of 0.357, outperforming the best alternatives (MAE = 0.594, PCC = 0.253). Detailed results for these marker genes are provided in the supplementary material. These results collectively highlight HiFusion’s robust and consistent superiority across diverse gene targets.


Figure~\ref{fig:fig3} illustrates the predicted spatial distributions of ERBB2, KRT19, and CD74 in three WSI samples, along with corresponding MAE and PCC scores for each model. Notably, HiFusion not only achieves the highest quantitative agreement with the ground truth, but also provides visually accurate localization of high-expression regions (highlighted by brighter colors), demonstrating its robustness in capturing complex spatial gene expression patterns.



\section{Discussion and Conclusion}\label{sec:conclusion}


We propose HiFusion, a novel framework for spatial gene expression prediction from whole-slide images. To overcome the limitations of coarse spot-level modeling and insufficient contextual integration, HiFusion combines hierarchical intra-spot modeling with context-aware cross-scale fusion, effectively capturing multiscale spatial and biological heterogeneity. The HISM module extracts fine-grained features across multiple resolutions, reinforced by a feature alignment loss to ensure semantic consistency. The CCF module further enhances representation through residual cross-attention, dynamically integrating intra-spot features with neighboring regional context. Extensive experiments demonstrate that HiFusion consistently outperforms competing methods and generalizes robustly across both 2D slide-wise cross-validation and 3D sample-specific evaluation. Importantly, we found that the 3D learning paradigm achieves strong intra-patient generalization at minimal computational and labeling cost, underscoring its practical potential for clinical spatial transcriptomics. While HiFusion effectively incorporates regional context via a single-branch design, future work may explore more expressive and efficient mechanisms to extract biologically-relevant fine-grained features from tissue regions and integrate them more effectively with intra-spot representations.



\bigskip
\noindent 

\bibliography{aaai2026}

\appendix
\section*{Supplementary Material}
\addcontentsline{toc}{section}{Supplementary Material}

\subsection{Novelty Clarification}
While multi-scale modeling and attention-based fusion are established concepts, our novelty lies in their biologically motivated integration specifically tailored for spatial transcriptomics gene prediction. Unlike prior works such as TRIPLEX and ASIGN, which treat each spot as homogeneous and primarily fuse spot-level and regional/global information, HISM explicitly captures intra-spot heterogeneity. Our hierarchical design models biological structures from tissue-level organization down to subcellular patterns, enabling the encoder to learn fine-grained, gene-relevant representations that fixed windowing or coarse multi-scale schemes cannot capture.
The multi-scale pathway is not intended to maximize per-scale discriminability; instead, it is designed to expose the shared encoder to stable, high-resolution cues. By leveraging CNNs’ translational invariance, our alignment module enforces semantic consistency across scales, yielding robust fine-grained features critical for accurate gene prediction.
Our CCF module introduces a conceptually new mechanism for integrating tissue context. Regional features serve as Queries, allowing macro-level tissue structures to guide the weighting of intra-spot representations. This attention-based fusion enables context-sensitive regulation rather than global averaging, reflecting biological relationships between tissue architecture and local molecular variation. The combination of intra-spot alignment and tissue-level contextual fusion has not been explored in prior histology-to-gene frameworks.

Experimentally, we are the first to evaluate representative baselines (ST-Net to ASIGN) under a 3D sample-specific testing paradigm, demonstrating that one-shot intra-sample training achieves high precision while remaining affordable and efficient for clinical deployment. HiFusion consistently outperforms all baselines, including more complex Transformer-based methods (HisToGene, His2ST, EGN, TRIPLEX, ASIGN), across both datasets and evaluation settings, and captures more accurate spatial expression patterns. Finally, our pipeline is architecture-agnostic, enabling flexible integration of alternative backbones as encoders without modifying the overall framework.

\subsection{Computational Cost Comparison}

Table~\ref{tab:tab4} presents a detailed comparison of computational costs between our proposed HiFusion framework and several baseline models, based on a single spot image from the ST-Data dataset. The table reports both the number of trainable parameters and the number of floating point operations (FLOPs).

For transformer-based models such as HisToGene and His2ST, where the input during inference is the collection of all spots within a whole slide, we divide the total FLOPs by the number of spots to estimate the per-spot computational cost. Similarly, TRIPLEX employs a global encoder that jointly processes all spot features from a slide; thus, we report the average FLOPs across all slide samples to represent its per-spot cost.

Although HiFusion introduces a trainable neighbor encoder, which leads to slightly higher FLOPs compared to some baselines, it demonstrates strong computational efficiency in two important aspects. First, HiFusion achieves state-of-the-art performance with substantially fewer parameters, avoiding the use of overly complex modules or transformer layers. Second, models such as EGN, TRIPLEX, and ASIGN require preprocessing steps, such as retrieving similar samples or extracting global and regional embeddings, which are not included in the FLOPs calculation but still incur additional inference cost. In contrast, HiFusion operates entirely without preprocessing, enabling highly efficient inference on unseen samples. These advantages collectively highlight the practicality and scalability of HiFusion, particularly in real-world deployment scenarios where both accuracy and computational efficiency are essential.

\begin{table}[t]
\centering
\begin{tabular}{c|cc}
\toprule
\textbf{Method} & \textbf{Param(M)} $\downarrow$ & \textbf{FLOPs(G)} $\downarrow$ \\
\midrule
ST-Net                       & 7.21 & 2.90  \\
HisToGene                 & 188.991 & 0.19  \\
His2ST                 & 672.43 & 1.49 \\
EGN                 & 44.11 & 1.88  \\
TRIPLEX           & 24.57 & 4.08  \\
ASIGN           & 25.39 & 3.00 \\
HiFusion (Ours)           & 17.39 & 9.18  \\
\bottomrule
\end{tabular}
\caption{Computational cost comparison}
\label{tab:tab4}
\end{table}

\subsection{Ablation Study on HiFusion Components}
We conduct a series of ablation experiments to evaluate the contribution of each component in \textit{HiFusion}, with quantitative results reported in Table~\ref{tab:tab8}. 

Removing the region branch, retaining only the HISM module, leads to a substantial decline in performance, confirming the importance of incorporating broader tissue context. Replacing the cross-attention mechanism in the CCF module with simple additive fusion also degrades performance, underscoring the necessity of attention-based context integration.

We further examine two variants that reverse the assignment of queries and keys. In the Q/K Reversed (CCF) variant, the inversion occurs within the CCF module: the globally averaged intra-spot representation functions as the query, whereas region features, pooled into a $k^2 \times d$ representation with $k=2$, serve as the keys and values. In contrast, the {Q/K Reversed (Input) variant performs the swap at the input stage, where HISM is applied to the region input so that region-derived features become the keys and values while the spot features act as the query.

Both variants lead to clear performance degradation, validating our original query–key formulation. These findings further support the biological prior that macro-scale tissue architecture modulates local molecular variation and should therefore guide intra-spot feature refinement.


\begin{table}[t]
\centering
\begin{tabular}{l|ccc}
\toprule
\textbf{Variant} & \textbf{MSE} $\downarrow$ & \textbf{MAE} $\downarrow$ & \textbf{PCC} $\uparrow$ \\
\midrule
HiFusion (Full)          & 0.5459 & 0.5699 & 0.4961 \\
w/o Region Branch           & 0.9265 & 0.7470 & 0.3870 \\
CCF (Additive)         & 0.6542 & 0.6141 & 0.4297 \\
Q/K Reversed (CCF)        & 0.5690 & 0.5806 & 0.4745 \\
Q/K Reversed (Input)      & 0.5586 & 0.5768 & 0.4771 \\
\bottomrule
\end{tabular}
\caption{Ablation analysis demonstrating the contribution of each HiFusion component on HER2.}
\label{tab:tab8}
\end{table}

\subsection{Additional Ablation Study on ST-Data}
We present additional ablation studies on the ST-Data dataset, with detailed results provided in Table~\ref{tab:tab5} and Figure~\ref{fig:fig4}. These experiments, excluded from the main manuscript due to space limitations, are designed to systematically evaluate the contribution of key architectural components within the proposed HiFusion framework. In particular, we assess the effects of varying image decomposition levels, the number of spot tokens, and the size of neighboring image patches on the accuracy of gene expression prediction.

The observed trends are highly consistent with those reported on the HER2 dataset, reinforcing the robustness and generalizability of our framework across different datasets. This consistency underscores the complementary roles of each individual module and their configurations, demonstrating that the full integration of these components is essential for achieving optimal performance. Collectively, these findings highlight the importance of hierarchical intra-spot modeling and context-aware cross-scale fusion in enhancing predictive accuracy in spatial gene expression tasks.

\begin{table}[t]
\centering
\begin{tabular}{l|ccc}
\toprule
\textbf{Level Combination} & \textbf{MSE} $\downarrow$ & \textbf{MAE} $\downarrow$ & \textbf{PCC} $\uparrow$ \\
\midrule
1×1                       & 0.5166 & 0.5589 & 0.5554 \\
1×1 + 2×2                 & 0.5135 & 0.5587 & 0.5595 \\
1×1 + 4×4                 & 0.5136 & 0.5587 & 0.5563 \\
1×1 + 7×7                 & 0.5148 & 0.5587 & 0.5554 \\
1×1 + 2×2 + 4×4           & 0.5143 & 0.5592 & 0.5592 \\
1×1 + 2×2 + 7×7           & \textbf{0.5095} & \textbf{0.5557} & \textbf{0.5613} \\
1×1 + 4×4 + 7×7           & 0.5182 & 0.5603 & 0.5512 \\
1×1 + 2×2 + 4×4 + 7×7     & 0.5173 & 0.5588 & 0.5593 \\
\bottomrule
\end{tabular}
\caption{Ablation study for image decomposition levels}
\label{tab:tab5}
\end{table}

\begin{figure}[t]
  \centering

  \begin{subfigure}[b]{0.9\linewidth}
    \centering
    \includegraphics[width=\linewidth]{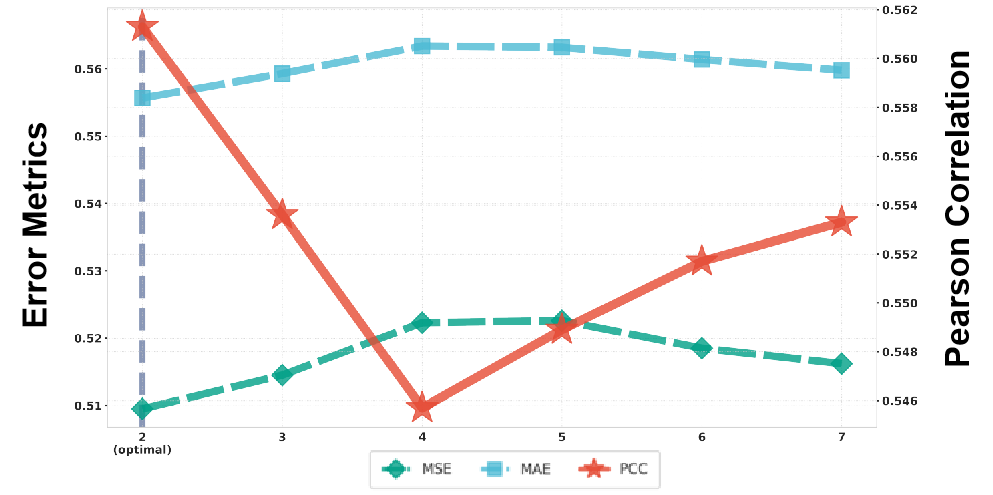}
    \caption{Spot Token Number $(k,k)$}
    \label{fig:token_num}
  \end{subfigure}

  \begin{subfigure}[b]{0.9\linewidth}
    \centering
    \includegraphics[width=\linewidth]{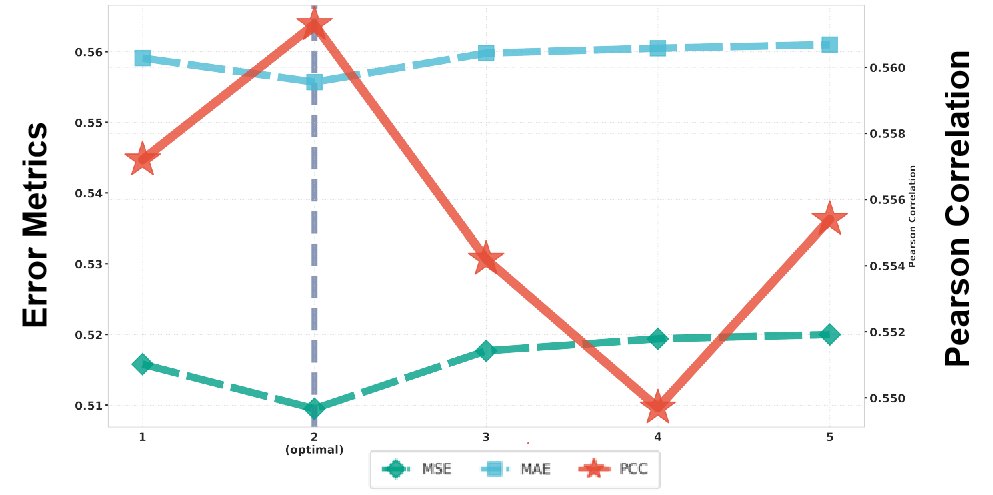}
    \caption{Neighbor Patch Size $(224 \times N,224 \times N)$}
    \label{fig:neighbor_size}
  \end{subfigure}

  \caption{Ablation study for (a) spot token number and (b) neighbor patch size.}
  \label{fig:fig4}
\end{figure}




\section{Comparison of 2D and 3D Training Paradigms}
\begin{table}[t]
\centering
\small
\label{tab:ablation_fa}
\setlength{\tabcolsep}{3pt} 
\begin{tabular}{c|c c c|c c c}
\hline
\multirow{2}{*}{\textbf{Method}} & \multicolumn{3}{c|}{\textbf{HER2}} & \multicolumn{3}{c}{\textbf{ST}} \\
\cmidrule(lr){2-4} \cmidrule(lr){5-7} 
& MSE & MAE & PCC & MSE & MAE & PCC \\
\midrule
HiFusion & 0.5543 & 0.5717 & 0.4899 
    & 0.5106  & 0.5563  & 0.5617  \\
HiFusion (3D) & \textbf{0.2846} & \textbf{0.4205} &\textbf{0.7492} 
    & \textbf{0.2711} & \textbf{0.4102} & \textbf{0.7838} \\
\hline
\end{tabular}
\caption{Performance comparison of HiFusion under 2D slide-wise and 3D sample-specific training on the HER2 and ST-Data datasets. The first slide of each patient is excluded from the 2D test set to match the 3D evaluation protocol.}
\label{tab:tab7}
\end{table}
\begin{figure}[t]
  \centering
  \includegraphics[width=\linewidth]{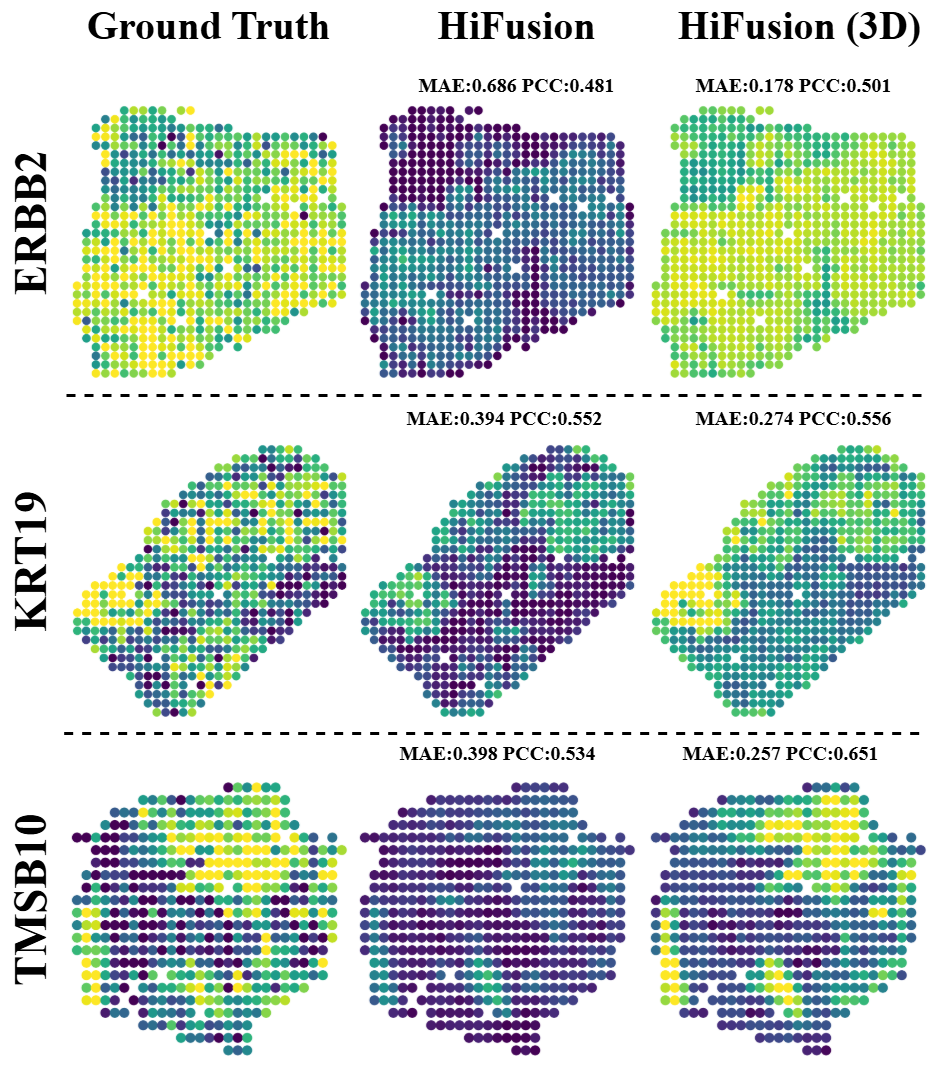} 
  \caption{Predicted spatial expression of ERBB2, KRT19, and TMSB10 on three representative samples from the HER2 dataset. HiFusion and HiFusion (3D) are compared with the ground truth. Brighter regions indicate higher gene expression. HiFusion (3D) shows better visual and quantitative alignment.}
  \label{fig:fig5}
\end{figure}
\begin{table*}[t]
\centering
\begin{tabular}{c|ccc|ccc|ccc|ccc}
\toprule
\multicolumn{1}{c|}{}                         & \multicolumn{3}{c|}{ERBB2}                                                   & \multicolumn{3}{c|}{KRT19}                                                   & \multicolumn{3}{c|}{CD74}                                                    & \multicolumn{3}{c}{TMSB10}                                                  \\
\cmidrule(lr){2-4} \cmidrule(lr){5-7} 
\cmidrule(lr){8-10} \cmidrule(lr){10-13}
\multicolumn{1}{c|}{\multirow{-2}{*}{Method}} & \multicolumn{1}{c}{MSE} & \multicolumn{1}{c}{MAE} & \multicolumn{1}{c|}{PCC} & \multicolumn{1}{c}{MSE} & \multicolumn{1}{c}{MAE} & \multicolumn{1}{c|}{PCC} & \multicolumn{1}{c}{MSE} & \multicolumn{1}{c}{MAE} & \multicolumn{1}{c|}{PCC} & \multicolumn{1}{c}{MSE} & \multicolumn{1}{c}{MAE} & \multicolumn{1}{c}{PCC} \\
\midrule
STNet  & 1.174                   & 0.851                   & 0.375                   & 0.370                   & 0.490                   & 0.158                   & 0.614                   & 0.615                   & \underline{0.253}                   & 0.295                   & 0.424                   & 0.227                   \\

HisToGene    & 1.067                   & 0.865                   & 0.280                   & 0.425                   & 0.523                   & 0.088                   & 0.835                   & 0.730                   & 0.092                   & 0.266                   & 0.412                   & 0.163                   \\

His2ST    & 1.021                   & 0.871                   & -0.080                  & 0.380                   & 0.493                   & 0.017                   & 0.666                   & 0.632                   & -0.034                  & 0.255                   & 0.385                   & 0.035                   \\

EGN      & \underline{0.797}                   & \underline{0.754}                   & 0.398                   & 0.471                   & 0.563                   & 0.111                   & \textbf{0.536}                   & \underline{0.594}                   & 0.167                   & \underline{0.229}                   & \underline{0.358}                   & 0.270                   \\
 
TRIPLEX      & 0.896                   & 0.810                   & \underline{0.481}                   & 0.394                   & 0.514                   & \underline{0.201}                   & 0.649                   & 0.653                   & 0.238                   & 0.310                   & 0.437                   & \underline{0.316}                   \\

ASIGN      & 1.085                   & 0.880                   & -0.031                  & \underline{0.332}                   & \underline{0.450}                   & 0.031                   & 0.705                   & 0.654                   & 0.017                   & 0.473                   & 0.490                   & -0.023                  \\

HiFusion (Ours)                                          & \textbf{0.720}          & \textbf{0.711}          & \textbf{0.518}          & \textbf{0.309}          & \textbf{0.446}          & \textbf{0.230}          & \underline{0.555}          & \textbf{0.584}          & \textbf{0.357}          & \textbf{0.186}          & \textbf{0.313}          & \textbf{0.355}      \\
\bottomrule
\end{tabular}
\caption{Performance comparison on individual cancer marker genes (ERBB2, KRT19, CD74, and TMSB10) across different methods. Best and second-best results are shown in bold and underlined, respectively.}
\label{tab:tab6}
\end{table*}
In this section, we compare the quantitative and visual performance of HiFusion under two training paradigms: 2D slide-wise training and 3D sample-specific training (denoted as HiFusion (3D)). We evaluate the predicted spatial expression of three cancer marker genes (ERBB2, KRT19, and TMSB10) in both settings. As shown in Figure~\ref{fig:fig5}, HiFusion (3D) produces gene expression patterns that better align with the ground truth, with lower MAE and higher PCC, and more accurately highlights regions of high expression (brighter areas), indicating superior spatial fidelity. For a fair comparison, both paradigms are evaluated on the HER2 and ST-Data datasets (Table~\ref{tab:tab7}). For HiFusion (2D), we exclude the first annotated slide sample from each patient in the test set, aligning the evaluation protocol with that of HiFusion (3D), where the first slide of each patient is consistently used for training. The results show that HiFusion (3D) consistently outperforms its 2D counterpart across both datasets, confirming the advantage of sample-specific learning in capturing intra-patient consistency and improving generalization.

These findings highlight two key insights. First, with annotations from only a single intra-patient slide, the model can achieve strong generalization within the same patient, demonstrating the effectiveness of intra-patient learning. Second, the substantial domain gap and variability across patients limit the robustness of models trained on multiple inter-patient samples, thereby hindering their ability to generalize to unseen patients. Taken together, these results highlight 3D sample-specific learning as a promising future direction, offering improved generalizability and reduced annotation costs. Beyond this, while HiFusion effectively integrates regional context via a single-branch design, future work may explore more expressive and efficient strategies to capture biologically meaningful fine-grained features and better fuse them with intra-spot representations.

\subsection{Additional Experiments on Cancer Marker Genes}

In this section, we further evaluate the gene-level prediction capabilities of the proposed method under the cross-validation setting on the HER2 dataset, focusing on four clinically important cancer marker genes: ERBB2 \cite{mehta2014co}, KRT19 \cite{saha2017krt19}, CD74 \cite{su2017biological}, and TMSB10 \cite{yan2021tmsb10}. These genes were selected due to their established relevance in cancer diagnosis and progression.

As reported in Table~\ref{tab:tab6}, our method consistently outperforms the baseline models across all four genes in terms of MSE, MAE, and PCC. These results demonstrate that HiFusion is not only effective at global expression prediction but also highly accurate in modeling the spatial expression patterns of key oncogenes. This highlights the potential of our framework for downstream clinical and translational research applications.

\subsection{Selected Genes}

Following the methodology outlined in ST-Net \cite{he2020integrating}, we selected the top 250 genes with the highest average expression for prediction. Figure ~\ref{fig:fig6} provides details of the selected genes and the corresponding codes for each dataset.
\begin{figure*}[t]
\centering
\includegraphics[width=\linewidth]{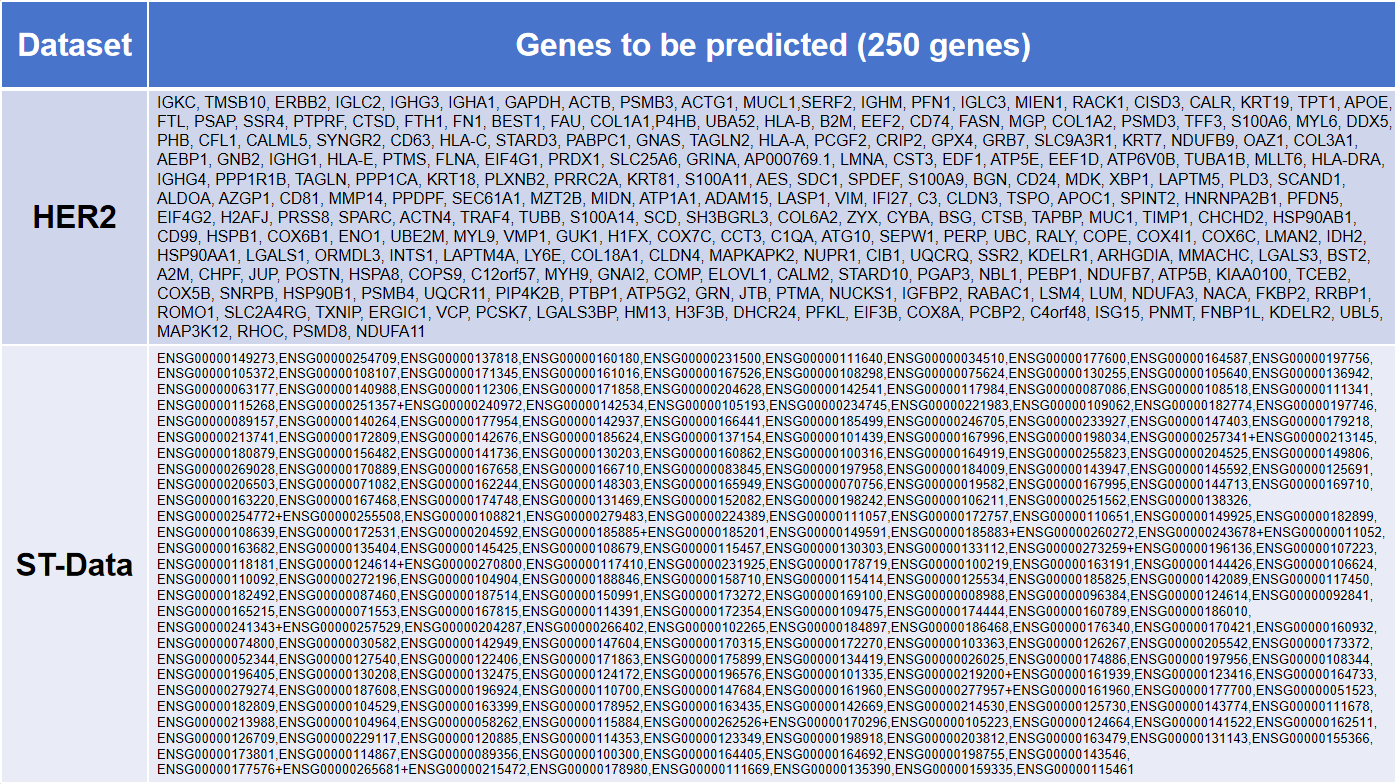} 
\caption{Genes selection in each public dataset. This figure showcases the top 250 genes with the highest expression levels for each public dataset utilized in this paper.}
\label{fig:fig6}
\end{figure*}



\end{document}